# Deep Neural Network Approach for Annual Luminance Simulations

Yue Liu[1], Alex Colburn[2], Mehlika Inanici[1]

[1]*College of Built Environment, University of Washington, Seattle, WA, USA*
[2]*Allen School of Computer Science and Engineering, University of Washington, Seattle, WA, USA*

*Corresponding author:* Yue Liu
*Email*: yueliu@uw.edu

Annual luminance maps provide meaningful evaluations for occupants' visual comfort, preferences, and perception. However, acquiring long-term luminance maps require labor-intensive and time-consuming simulations or impracticable long-term field measurements. This paper presents a novel data-driven machine learning approach that makes annual luminance-based evaluations more efficient and accessible. The methodology is based on predicting the annual luminance maps from a limited number of point-in-time high dynamic range imagery by utilizing a deep neural network (**DNN**). Panoramic views (with 360° horizontal and 180° vertical field of view) are utilized, as they can be post-processed to study multiple view directions. Indoor luminance and outdoor sky maps are provided as input along with sun position and solar radiation data. The output is predictions of indoor luminance maps under novel sky conditions. A thorough sensitivity analysis is performed to develop guidelines for determining the minimum and the optimum data collection periods, and the minimum number of imageries needed for generating accurate luminance maps. The proposed DNN model can faithfully predict high-quality annual panoramic luminance maps from one of the three options within 30 minutes training time: i) point-in-time luminance imagery spanning 5% of the year, when evenly distributed during daylight hours, ii) one-month hourly imagery generated or collected continuously during daylight hours around the equinoxes (approximately 8% of the year); or iii) 9 days of hourly data collected around the spring equinox, summer and winter solstices (2.5% of the year) all suffice to predict the luminance maps for the rest of the year. The DNN predicted high-quality panoramas are validated against Radiance (RPICT) renderings using a series of quantitative and qualitative metrics. The most efficient predictions are achieved with 9 days of hourly data collected around the spring equinox, summer and winter solstices. The results clearly show that practitioners and researchers can efficiently incorporate long-term luminance-based metrics over multiple view directions into the design and research processes using the proposed DNN workflow.

## Introduction

Light entering the human ocular system results in image forming (visual) and non-image forming (circadian) response. The visual impact of lighting can be best studied through image processing. Per-pixel luminance values and the resulting luminance distribution patterns in physically accurate imagery determine the human visual comfort and preference (Wienold and Christoffersen 2006, Jakubiec and Reinhart 2012, Suk et al. 2013, Konis 2014, van den Wymelenberg and Inanici 2016), task visibility, and visual appearance and perception (Rockcastle et al. 2017).

Historically, lighting measurements and recommendations were done with illuminance measurements. Illuminance meters cost less; therefore, they were more readily available to researchers and practitioners. High Dynamic Range (**HDR**) photography technique (Debevec and Malik 1997) allowed professionals to measure luminance at a per-pixel scale (Inanici 2006), and this ability fostered the development of several human-centric lighting metrics in the last two decades. The luminance maps are useful to study the human response to the environment, whereas illuminance maps are mainly useful to determine potential energy savings. The goal in any successful daylighting design is to support human health, satisfaction, and productivity along with significant energy savings. As a result, it would stand to reason to expect that luminance maps in lighting simulations to be computed, at least in the same frequency, as illuminance values. However, similar to measurements, grid-based illuminance simulations are less costly than per-pixel luminance maps due to the resolution of the calculation points. Climate-based (annual) illuminance simulation techniques have been developed (Nabil and Mardaljevic 2005, Reinhart et al. 2006) and quickly adopted in design and research communities. Climate-based luminance simulation techniques (Ward et al. 2011, McNeil et al. 2013, Lee et al. 2018) are not widely adopted, as they have steep learning curves and long simulation time.

Recognizing the need to improve the availability and accessibility of annual luminance-based lighting simulations, the goal of this research is to develop a deep neural network framework to accelerate long-term per-pixel luminance predictions by generating annual luminance maps from a small subset of HDR images. The deep learning algorithm i) learns the parameter values for a high dimensionality regression function, ii) by optimizing over a training set of physically accurate luminance and sky image samples, and features such as the position of the pixels, azimuth and altitude angles of the sun, and direct normal and diffuse horizontal irradiances, iii) extracts relevant luminance-based characteristics, iv) improves the parameter values in response to prediction errors, v) and predicts per-pixel luminance images under novel sky conditions. In technical terms, the deep learning algorithm computes the radiance regression function, which is a non-linear mapping from local and contextual attributes of surface points to their luminance values. The specific objectives of this research are:

- To develop a deep learning framework to conduct architectural luminance predictions;
- To demonstrate a robust workflow that accurately and efficiently predicts annual panoramic luminance maps from a small number of HDR images;
- To evaluate the results through a number of quantitative and qualitative metrics;
- To perform sensitivity analyses to determine the minimum number of HDR images necessary to accurately generate annual luminance data. Sample data size informs both the optimum and minimum data collection period and the minimum sample size; and
- To discuss the impact of the weather and seasonal changes on the accuracy of the outcome, given a fixed data collection period.

## Background on Accelerating Annual Daylighting Calculations

Daylight Coefficient (**DC**) method was originally developed by Tregenza and Waters (1983). It is based on dividing the sky into a finite number of discreet patches. DCs are calculated at measurement points as the normalized lighting values from each sky patch (which does not change as long as the site, building geometry, and material properties are not modified). The lighting quantity (illuminance or luminance) at a sensor point is calculated by adding the contributions of individual sky patches that are computed by scaling the DCs with the actual sky luminance of each sky patch at a given date and time. Bundled simulations are performed to pre-compute the resulting effect of each sky patch for each sensor point; the number of sky patches determines the number of simulations. The original Tregenza sky division has 145 patches, so 145 simulations are performed, and matrix calculations are done to predict approximately 4000 daylight hours in a given year. Luminance based DC approaches require multiple phases to model the direct and diffuse components of daylight; and finer resolutions (5185) are necessary to adequately model the sun penetration patterns (Ward et al. 2011, McNeil et al. 2013, Lee et al. 2018). These methodologies accelerate annual luminance calculations, but they still require a substantial amount of computing time.

Recent accelerating methodologies utilize parallel computing to trace multiple primary rays simultaneously on a graphics process unit (**GPU**) (Zuo et al. 2011, Jones and Reinhart 2017). Due to its highly parallel structure, a modern GPU can efficiently process a large block of data, which is not achievable by a general-purpose central processing unit. Another approach is to utilize machine learning to create predictive models from observed data, bypassing the expensive computations involved with physically-based simulations. Lighting researchers have investigated machine learning, and specifically neural networks, for indoor illuminance predictions, with applications in the operation of electric lighting systems, energy estimations, and optimizations of shading systems. Conventional feedforward neural networks are utilized to estimate illuminance levels under daylight (Kazanasmaz et al. 2009, Ahmad et al. 2017, Katsanou et al. 2019) and illuminance distributions under electric lighting (Sahin et al. 2016) from input parameters related to light sources and room settings. Although inspiring, illuminance predictions, in general, are less computationally intensive compared with luminance predictions due to lower resolutions of calculation points, and lack of dependency on view directions. Annual illuminance predictions can be achieved through DC simulation practices without significant simulation time. Statistical methods have been developed previously for per-pixel luminance predictions (Inanici 2013), but to the best of the authors' knowledge, there is no previous daylighting research that utilizes machine learning to predict long-term luminance maps.

## Background on Artificial Neural Networks

An artificial neural network (**ANN**) (McCulloch and Pitts, 1943) is a powerful machine learning technique for modeling complex systems. ANNs are composed of neuron-like connected computing nodes, organized into hierarchic layers from input layers, hidden layers, to output layers. Each pair of connected nodes are assigned a weight parameter, which controls the degree of influence a node has on its connected pair. Each node has a bias parameter which defines how much the output of the transformation at the node is biased in the absence of any input. An ANN finds the relationships between inputs and outputs through a learning process. The learning process involves multiple steps: 1) input data passes through the input layers towards the output layers; 2) at the output layer, the predicted results are calculated and compared with the ground truth using an error metric, termed a loss function; 3) the weight parameters are updated according to the calculated errors; 4) the previous steps are repeated several times (epochs) until the error between the prediction and ground truth is minimized. After the model is trained, it can be used for predictions, given new input data. A DNN consists of more hidden layers between the input and output layers in comparison to a shallow ANN and can learn more complicated non-linear relationships.

While only a few previous studies existed in daylighting research, machine learning has been increasingly employed in closely related problems, such as rendering and appearance synthesis, in computer vision and graphics. Rendering a scene under novel lighting conditions has been a long-lasting research question in computer vision and graphics fields with applications in virtual reality, augmented reality, and visual effects. Machine learning has been investigated for image de-noising (Bako et al. 2017, Chaitanya et al. 2017). By creating high-quality images from noisy renderings with a reduced sample rate, the method has great potentials in accelerating physically-based rendering processes.

Ren et al. (2015) use a neural network model to learn the non-linear mapping of per-pixel RGB intensities from local and contextual attributes of surface points. The model is trained with a small subset (hundreds) of images of the scene to create images under new lighting conditions. The developed method has been tested with captures of the indoor scenes lit by point light sources. Gardner et al. (2017) use two individual deep neural networks to predict the positions and the intensities of light sources of indoor scenes, from a single low dynamic range photograph. The estimated light sources are used to relight the virtual objects into the scene and composited into photographs. However, with the limited input information, this study aims at producing results with a plausible visual appearance instead of physical accuracy. The light sources are blurred, and their intensities are predicted less accurately than their locations. Xu et al. (2018) develop a neural network model to synthesize scene appearance under novel illuminations from five images. The model is trained on a large synthetic dataset that consists of objects with various shapes and reflectances to learn a complex, non-linear function across scenes. By leveraging a single deep neural network model than several shallow neural networks, the computing requirement is reduced. This method is limited to relighting the objects rather than scenes and does not model the interreflections and shadows. The study has a limited applicable luminance range, less than the wide dynamic range of daylit scenes.

Machine learning has also been used to approximate high luminance range sky models from images. Satilmis et al. (2016) developed a neural network model to predict HDR skies from a sparse set of HDR captures of various sun positions and trained the network with illumination/image pairs. After training, given the parameters of the sun position, view direction, and the sky turbidity level, the model can successfully predict an arbitrary HDR sky (excluding the sun pixels). Hold-Geoffroy et al. (2017) proposed a deep learning-based method to predict outdoor illumination represented by sky models. In their study, a Convolutional Neural Network (**CNN**) model is trained to predict the three parameters for reconstructing the Hošek-Wilkie sky model (2012) from a single Low Dynamic Range (LDR) image. Although these studies are useful to demonstrate the applicability of machine learning on daylighting predictions, they focus on modeling outdoor skies; therefore, they do not include the complex external and internal reflections that occur indoors along with the contributions of the sun and the sky.

## Methodology

### The Setting

The setting is an open plan office with a south-facing side window, located in Seattle (47.6°N, 122.3°W). The dimension of the room is 6m (width) by 14m (depth) and 4.5m (height). The camera is placed at the front center of the room at the eye level of 1.6m from the floor to correspond to a standing human's perspective (Figure 1). All images are rendered as equi-rectangular (spherical) panoramas from a single viewpoint using the Radiance rendering engine (Ward, 1994). Unlike the perspective or fisheye projections, panoramas allow full degree-of-freedom in camera roll, pitch, and yaw that can account for an occupant's changing view and gaze direction in a given environment (Figure 2). The rendered images in the database have a broad luminance range ($0$-$10^{+8.3}$cd/m$^2$) as the direct sun may appear in the field of view. The materials for walls, ceiling, floors, and outdoor ground have reflectance values of 50%, 80%, 20%, and 20%, respectively.

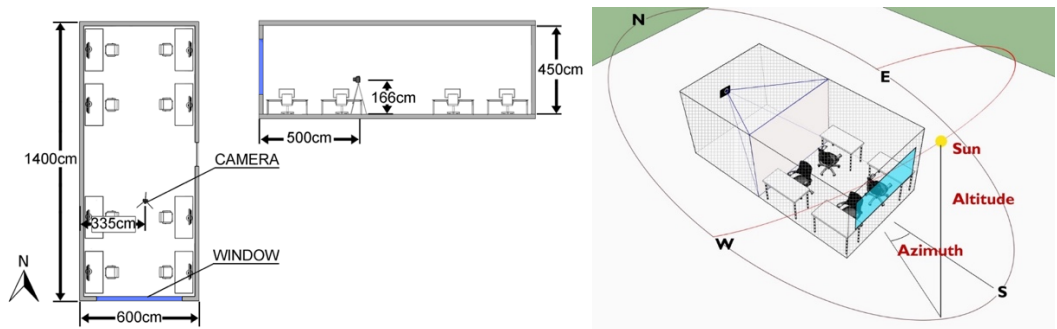

Figure 1. Test scene used in Radiance simulations.

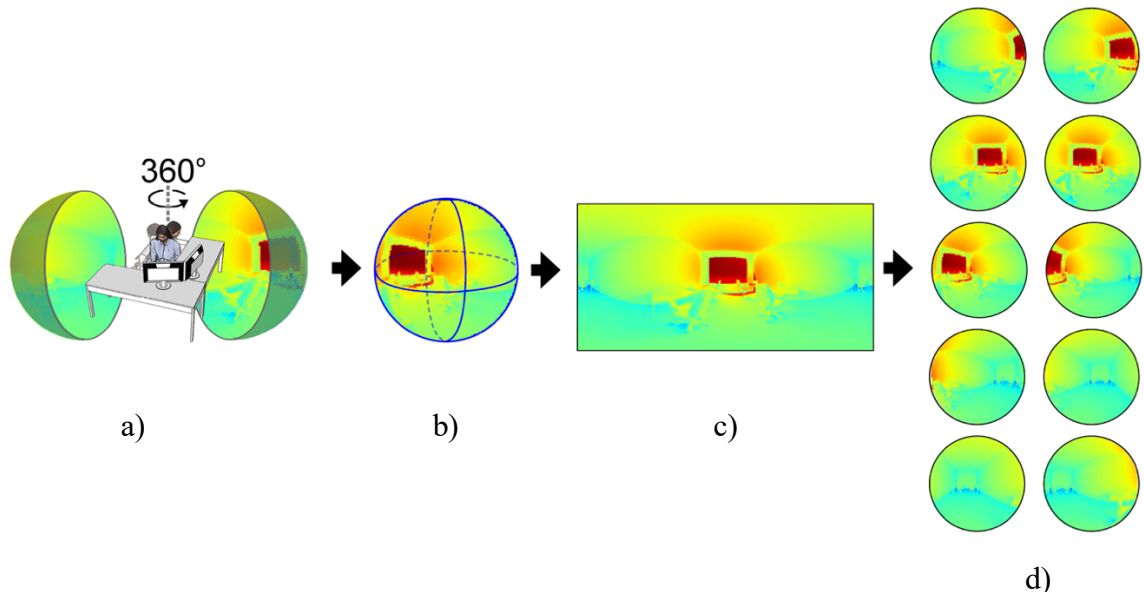

Figure 2. a and b) Full 360° field of view models an occupant's visual experience and provides more complete evaluations than a single fixed perspective view; Radiance generated equi-rectangular panoramas are used as (c) inputs to the DNN and then converted to fisheye images (d) for visual comfort analysis.

Three sets of images are rendered using the Radiance *RPICT* method: 1) High-quality interior maps are rendered with 4 ambience bounces (-ab 4); 2) The low-quality interior sun patches are rendered with 0 ambient bounce (-ab 0); they share the other RPICT rendering parameters (-ps 2 -pt .05 -pj .9 -dj .7 -ds .15 - dt .05 -dc .75 -dr 3 -st .15 -aa .1 -ar 512 -ad 2048 -as 1024 -lr 8 -lw .005); 3) The sky maps are generated with the same rendering parameters as the interior maps. All data is generated with the Perez all-weather sky model (Perez et al. 1993) using the direct and diffuse irradiances from an EnergyPlus weather file. The panoramas are rendered with a pixel resolution of 1840×920 and downsized to 460×230 using Radiance *pfilt* for anti-aliasing. All sets of images are generated in 1-hour intervals from sunrise to sunset for the entire year, resulting in 4379 data samples. Generating such a large dataset is necessary for developing and testing the machine learning model. When applying the workflow into practice, the user only needs to provide a small amount of HDR images.

### *DNN Model*

A light transport model describes how light travels in a scene. In traditional Radiance renderings, a light transport model is computed using a backward ray-tracing technique using scene information such as light sources, geometries, and materials. In this study, A DNN model is used to estimate the light transport model ($M$) through learning the non-linear relationship between illumination conditions (input $I$) and the corresponding pixel luminance values (output $L$). The estimated light transport model ($M$) allows the scene to be rendered under new lighting conditions. Using this method, generating annual luminance maps is a three-step process that involves: 1) acquiring sparse samples of luminance maps; 2) using the sparse samples to train a DNN that estimates the light transport model, and; 3) predicting annual luminance maps using the DNN.

When creating a DNN model, there are several considerations: training data sets, loss functions, model training, network architecture, and evaluation of the model's performance.

### *Training Sets*

While the DNN network architecture, and loss functions are important to achieve high accuracy, selecting the training set is equally important. If the training set is not balanced, the network will not generalize to a wide variety of inputs. For example, random sampling over the sky condition parameters results in a biased training set that skews towards diffuse lighting conditions with little direct light in this setting. Training with random sampling produces a model that does not generalize to the full range of lighting conditions. K-means clustering (a clustering method which partitions the data into *k* relatively equal-sized groups) is used to select the training samples evenly distributed over the light domains (Figure 3) in order to cover various sun positions and sky conditions. Three different training set options have been evaluated:

- Training Set 1: point-in-time luminance imagery spanning 5% of the year when evenly distributed during daylight hours,
- Training Set 2: one-month hourly luminance imagery generated or collected continuously during daylight hours; or
- Training Set 3: 9 days of hourly data collected around the spring equinox, summer and winter solstices (2.5% of the year).

Training Set 1 is discussed here. The other two options are discussed in the next Section.

The generated images are divided into training, validation, and test sets. The model learns to estimate the light transport model from the data in the training set and evaluates its performance with the validation set during training. After the model is trained, its performance is evaluated using the test set. Note that there is no overlap between the test/validation dataset and the training set. Hence, all tests are performed for lighting conditions that have not been seen by the network before.

*Loss functions*

Loss functions are a mechanism to compare ground truth to the predicted output of a model. Choice of loss functions greatly impacts DNNs and their learning dynamics. The model is trained with a combination of two loss functions: $\text{Loss}(y,t) = \mathcal{L}_{MSE}(y,t) + \lambda \times \mathcal{L}_{RER}(y,t)$, where y is the network prediction, t is the ground truth, and $\lambda$ controls the relative weights between two functions. The first function calculates the L2 distance (mean square error, **MSE**) between the predicted and ground truth luminance $\mathcal{L}_{MSE}(y,t) = \frac{1}{n}\sum_{i=1}^{n} \omega_i (y_i - t_i)^2$, where $\omega_i$ is the solid angle for pixel i. Solid angle weight $\omega_i$ is added so that each pixel is appropriately weighted for spherical panoramic solid angle projection (Figure 4**Error! Reference source not found.**). A second loss of Relative Error Rate (**RER**), $\mathcal{L}_{RER}(y,t) = \omega_i \sqrt{\frac{\sum_i (y_i - t_i)^2}{\sum_i t_i^2}}$ is added to improve the model performance. The weight $\lambda$ is set to 10 as the result of a parameter search.

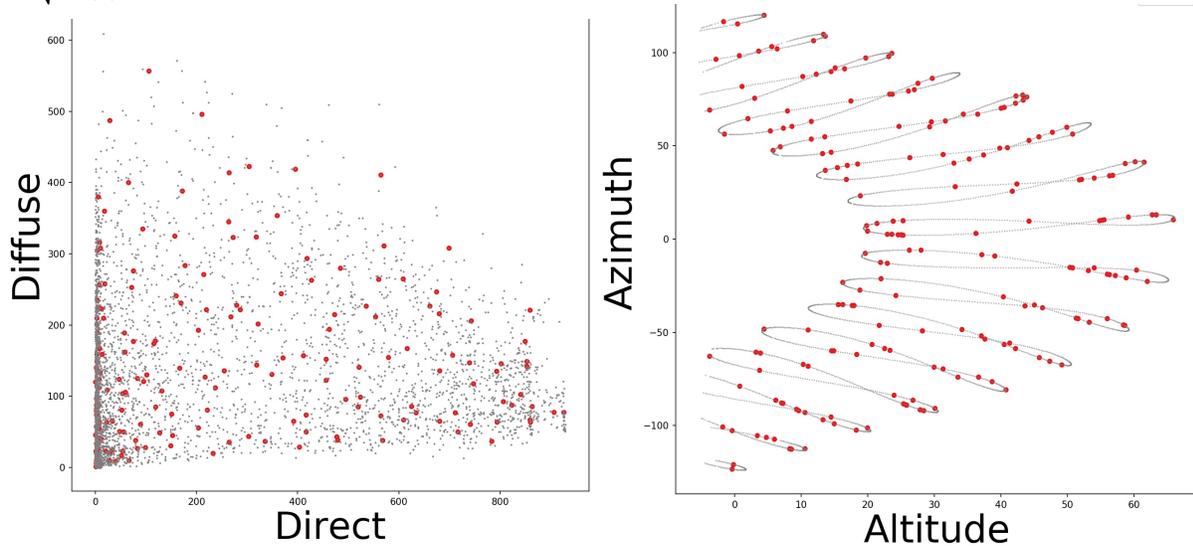

Figure 3. The K-means method is utilized to make the selected training samples well-distributed over the 4-dimensional light domain visualized in two 2-dimensional plots: (Left) distribution plot over sky condition parameters (direct and diffuse irradiances), and (Right) over sun position parameters (azimuth and altitude angle). Red dots represent the selected training samples while grey dots represent all data points. In this example, k= 200.

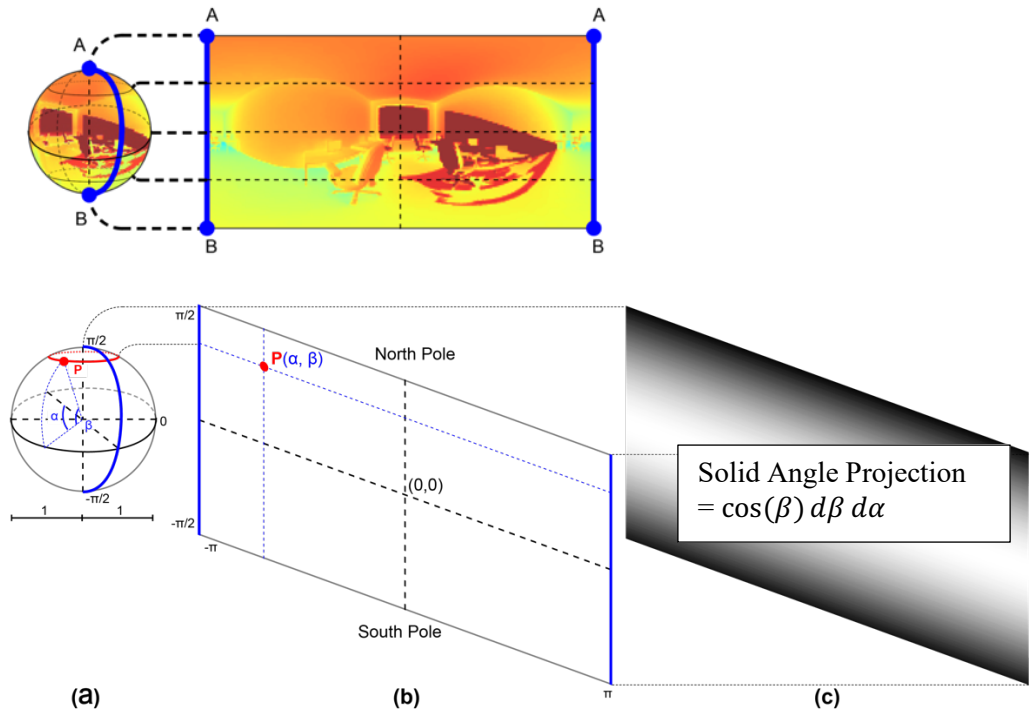

Figure 4. (Top) The equi-rectangular (spherical) panorama is mapped from a unit sphere. The north pole (A) and south pole (B) on the sphere become the top (AA) and bottom (BB) edges of the corresponding panorama. (Bottom) The image shows the solid-angle projection of each pixel in the panorama.

### Model Training

The model is trained with the ADAM optimizer (Kingma and Ba, 2015) with an initial learning rate of $10^{-3}$, reducing the learning rate by a factor of 2 once learning stagnates. Mini-batch learning starts with an initial batch size of $6 \times N$, where $N = \text{width} \times \text{height} \times 8$ is the total number of input parameters in one image, reducing the batch size by a factor of 2 after 30 epochs or if the learning rate is reduced to a threshold of $10^{-10}$. The minibatch size of $6 \times N$ is selected, so it fully utilizes the 11GB memory of the GPU. After a complete pass through the training set, the data is shuffled. The model is trained until the batch size is reduced to a minimum threshold of $1 \times N$ or the validation loss reaches $10^{-10}$. During the training process, only the model with a reduced error on the validation set is saved. Training takes roughly 30 minutes on an NVIDIA GeForce $^{1080Ti}$ GPU. At test time, predicting a high-quality panoramic luminance map takes 1/10 second.

### Model Architecture

The methodology is inspired by Ren et al. (2015). However, their method cannot be directly applied to architectural daylighting predictions due to the following restrictions: 1) A luminous environment lit by naturally occurring sun and the sky is much more complex than an environment illuminated by a single point light source. The possible luminance range of a daylit scene is orders of magnitude higher than of the scene lit by an electric light source. 2) Their method has a high computing requirement. They use an ensemble model created via a hierarchical clustering mechanism, which needs to be trained on a CPU cluster. After several rounds of developments and refinements, our machine learning model differs from Ren's in that it utilizes a single DNN rather than an ensemble model created via a hierarchical clustering mechanism. Tests show that the model achieves the same accuracy as Ren et al., and can be trained on a single machine rather than CPU cluster nodes.

We initially adopted a CNN approach. CNN is a deep learning algorithm, where multiple layers are trained in a robust manner (LeCun et al.1998). CNNs are similar to conventional ANNs in many ways but are designed for cases when the inputs are images, thus are commonly used in various computer vision research and applications. Unlike a conventional ANN, the neurons in layers of a CNN are arranged in three dimensions. Every layer transforms an input 3D volume to an output 3D volume through convolution operations. An input image can be seen as an input volume with width, height, and depth of 3 (RGB channels), which can be directly imported into CNNs.

The first task when designing a network architecture is to designate the input and output parameters. The input vector ($I$) of the CNN model is constructed from several parameters describing the illumination conditions, and the output vector ($L$) contains a sole value which is the pixel luminance in the given luminous environment. The input parameters include: 1) the sun location defined by sun altitude and azimuth angles (*al, az*); 2) the sky condition defined by direct and diffuse irradiances (*dir, dif*); and 3) the pixel location defined by the x and y coordinates (*px, py*). These six parameters describe the luminous environment within an indoor scene at a given date and time. The input parameters are augmented with the average luminance of a pixel (*avg*) to improve convergence (Ren et al. 2015), the pixel luminance of the sun patches (*sunpatch*) to aid the neural network in reconstructing sharp shadows and sun penetrations, and Perez sky luminance maps (*skymap*) to provide spatial information of the sky luminance distributions. Both sun patches and sky maps can be simulated quickly, thus have little impact on the simulation efficiency. The input feature vector includes 9 parameters: *px, py, al, az, dir, dif, avg, sunpatch, skymap.*

Figure 5 gives an overview of the research framework. The interior maps are used as ground truth, and the sky maps and sun patches are provided as input to the neural network training. The input parameters and the output luminance is preprocessed and normalized to the range of [0,1], with a normalization function x = $\frac{x - x_{max}}{x_{\max} - x_{min}}$. Normally, $x_{max}$ and $x_{min}$ should be selected as the highest and lowest value of the parameter x in the training set and applied to the validation and test set when evaluating the model. However, the training set only contains a limited number of images, which may not cover the entire luminance range in the database. Setting the $x_{min}$ and $x_{max}$ of the luminance parameters to be known luminance range of daylit scenes as 0 and 1.6 ×10$^{+9}$ cd/m$^2$ as the luminance of solar disc (Karandikar, 1955) provides a generic range for any given scene.

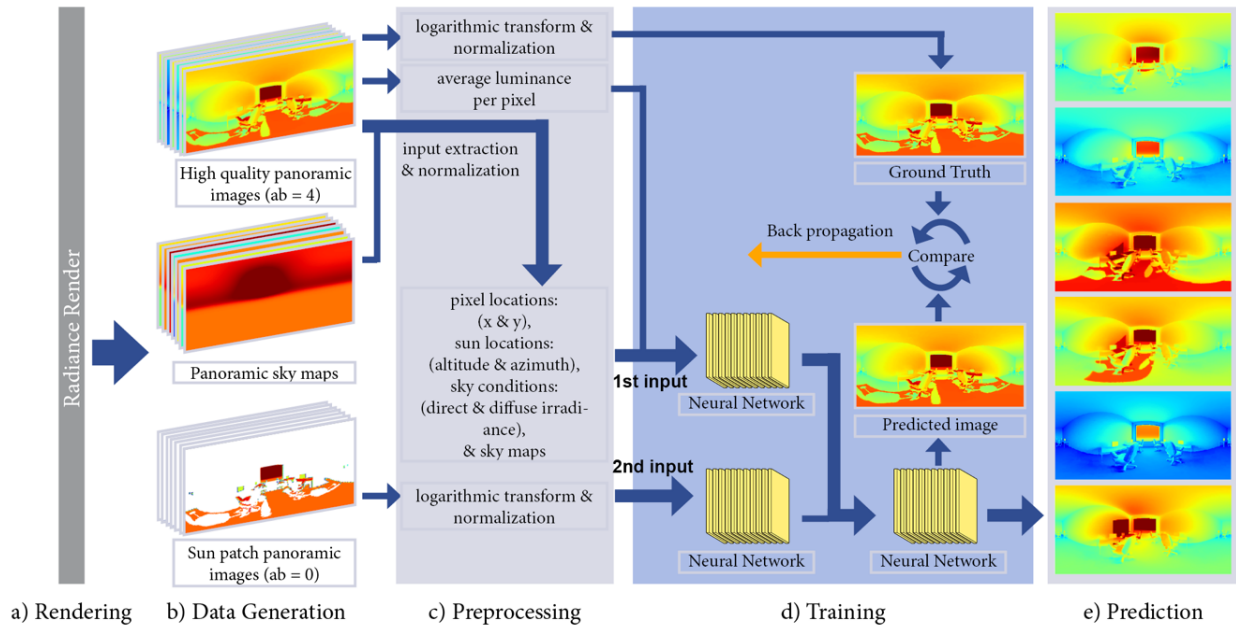

Figure 5. Overview of the research framework: a) A sparse data set is rendered using the Radiance simulation engine. b) Each data point includes three panoramas consisting of interior maps, sky maps, and sun patches. The interior maps are used as ground truth for the DNN training process, while the sky maps and the sun patches are used as the input to the DNN. c) All images are pre-processed and normalized. The sun patches and simulation parameters are combined to create the input for the DNN. d) The network is trained to predict the interior maps. e) The trained network can then be used to generate interior maps from novel sky maps, sun patches, and simulation parameters. Further long-term luminance-based analysis (e.g., glare and spatial contrast analysis) can be performed using these predicted interior panoramas.

Normalization reduces the variation in the parameters so that parameters of different scales and units contribute proportionately to the final results. However, normalizing the data can be problematic as the majority of the luminance values are low after scaled by the luminance of the solar disc (Figure 6(a)). To make the luminance values more evenly distributed over the range of [0, 1] after normalization, several additional preprocessing schemes are tested. Initially, gamma correction of 10 is applied to spread the distribution more evenly (Figure 6(b)). However, this scheme fails to highlight important image features, such as the edges and corners (Figure 7(b)). The optimum scheme combines multiple steps: A log10 transform is applied to spread the range of luminance more evenly (Figure 6(c)); the maps are normalized to the range of [0, 1]; a gamma correction of 1.5 is performed to shift the intensity distribution towards a mean value of ½ (Figure 6(d)). The model performance is significantly improved after the pre-processing with finer details (Figure 7(c)).

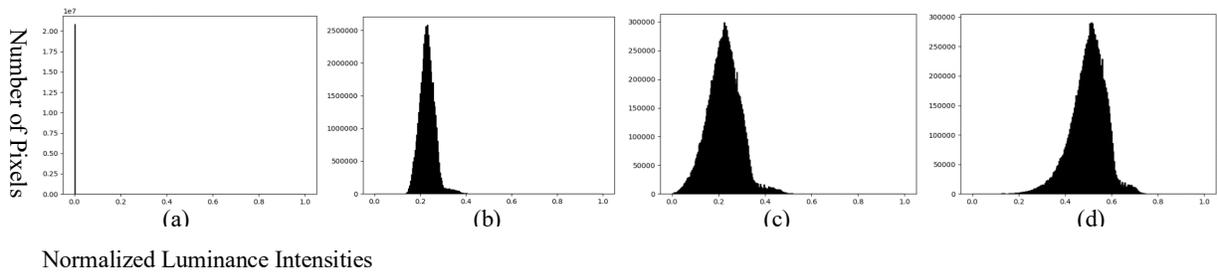

Figure 6. Diagrams show the distribution of the normalized luminance values with different preprocessing schemes: (a) the original, (b) gamma correction (gamma = 10), (c) $\log_{10}$ transform, and (d) $\log_{10}$ transform + gamma correction (gamma = 1.5); over the range of [0, 1]. Scheme (d) is selected as it spreads the range of luminance more evenly and shifts the intensity distribution towards a mean value of ½.

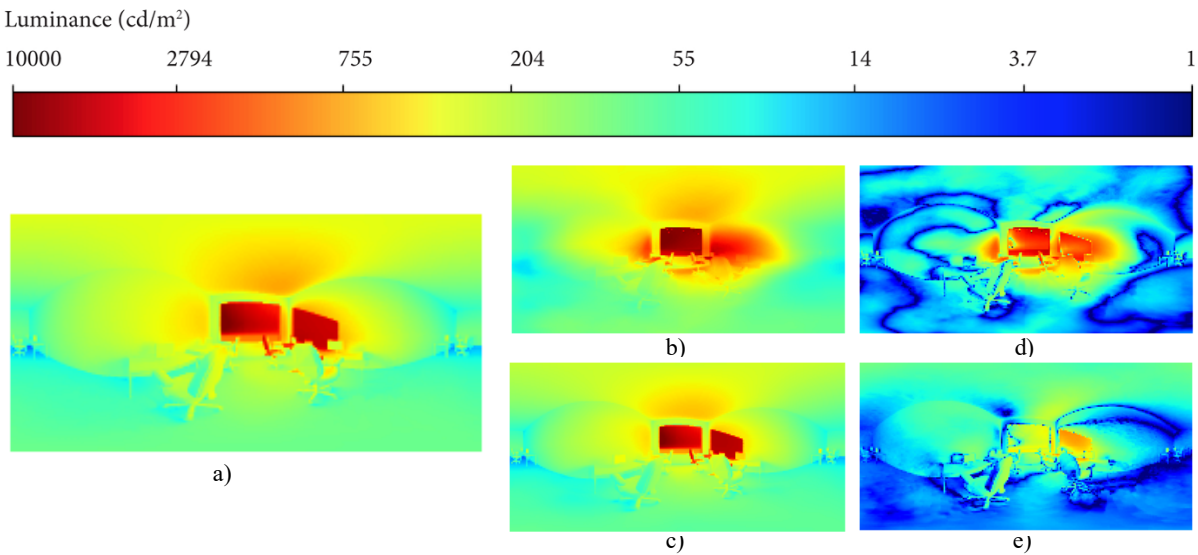

Figure 7. An exemplary test case (at 8am on October 25th with direct and diffuse irradiances of 476 and 77 W/m²) demonstrates the impact of data preprocessing schemes on prediction results. The ground truth luminance map as shown in (a). The predicted luminance maps of two processing schemes are compared: (b) gamma correction (gamma = 10) and (c) a combination of $\log_{10}$ transform and gamma correction (gamma = 1.5). (d) and (e) are errors maps for (b) and (c).

The initial network architecture divides the input features into three separate vectors to a three-input CNN: The first input branch, which encodes parameters describing global illumination, consists of four convolutional layers, each with 600 filters. The second and third input branches, which encode the direct sun illumination, consists of a single 200 filters convolutional layer. The two branches are concatenated to form an 800 filters layer, which is followed by a convolutional layer of 600 filters. The output layer is a convolutional layer with one filter that predicts luminance values. The Rectified Linear Units (ReLU) activation functions (Hahnloser et al. 2000, Nair and Hinton 2010) are applied to all layers. ReLUs have been shown to have good results for most of the feedforward neural networks (Goodfellow 2016). All convolutional layers use the same kernel size of 1 and stride of 1.

The CNN model exhibit difficulties in learning fine details under different lighting conditions. Since training the CNN requires the complete panorama input (460 × 230 × 9), small batch sizes are required during the training due to GPU memory limitations. Only a few lighting conditions are presented to the network in each batch step during training. The limited number of lighting conditions in each batch do not fully represent the overall distribution of the entire data set. Training does not converge as well as it could if each batch better represented the range of the data distribution. To overcome this limitation, the CNN architecture is converted into an equivalent multi-input deep dense network model architecture. A convolutional layer with a kernel size of 1 and stride of 1 is identical to a dense layer. The input data at each training step is sampled from various pixels and lighting conditions in the entire training set, rather than the few instances selected in each batch. Experiments show that the updated model and batch sampling strategy greatly reduces the prediction error (Table 1).

Table 1. The final developed scheme is compared with other schemes including 1) without sky panoramas as input; 2) utilized a CNN model instead of a dense network, and 3) with a different data preprocessing transform. The final scheme shows better performance on all metrics.

| Scheme | Preprocessing | Network Architecture | Sky Panoramas | $Log_{10}$ MSE | $Log_{10}$ RER | DGP MSE | RAMMG MSE |
|---|---|---|---|---|---|---|---|
| Final | $log_{10}$ + gamma 1.5 | Dense | Yes | 9.54e-03 | 4.85e-02 | 1.7e-04 | 29 |
| 1 | $log_{10}$ + gamma 1.5 | Dense | No | 9.78e-03 | 4.91e-02 | 2.0e-04 | 35 |
| 2 | $log_{10}$ + gamma 1.5 | CNN | Yes | 1.78e-02 | 5.61e-02 | 2.78e-04 | 41 |
| 3 | gamma 10 | Dense | Yes | 2.81e-02 | 8.31e-02 | 2.1e-03 | 151 |

The next key consideration is choosing the depth of the network and the width of each layer. The final network architecture is found via experimentation guided by monitoring the validation set error (Figure 8). Monitoring the training error is less meaningful as the error will always decrease with more trainable parameters, and the testing error should not be used for fine-tuning the parameters. The final developed network architecture divides the 9 input features into two separate vectors as input to a two-input deep network consisting of fully connected layers with ReLU activations. The first input branch, which encodes *px, py, al, az, dir, dif, avg*, and *skymap,* consists of four fully connected layers, each with 600 nodes. The second input branch, which encodes *sunpatch,* consists of a single 400 nodes fully connected layer. The first input branch encodes parameters describing global illumination, while the second input branch encodes direct sun illumination. The two branches are concatenated to form a 1000 node layer, which is followed by a fully connected layer of 600 nodes. The output layer is a single fully connected node that predicts luminance values.

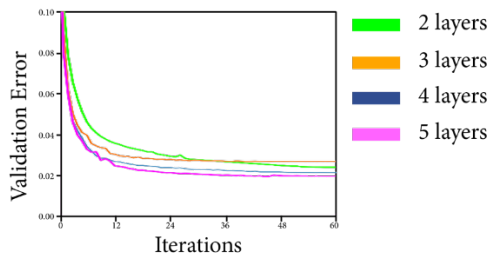

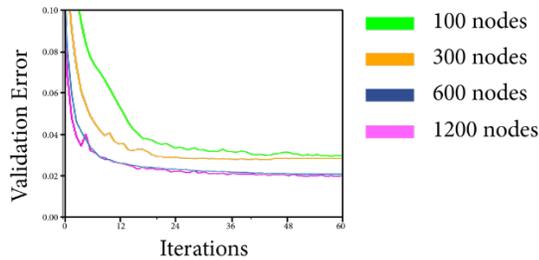

Figure 8. Diagrams show the comparison between the developed network model with different (Top) depths and (Bottom) width of each layer. The average validation error is measured after each iteration (i.e., a complete pass through the training set) as training proceeds. Convergence is attained after 28 to 30 passes. (Top) The model with a depth of 4 and 5 layers has the lowest validation error. 5-layers model has a slightly lower error, but considering the computational cost, 4-layers architecture is selected. (Bottom) The model with a width of 600 nodes in each layer has similarly low validation error as the 1200 nodes one, but with less computational cost.

## Results and Discussion

An initial sensitivity analysis is performed to study the impact of the number of training samples on the prediction accuracy and to find the minimum sample size. The training sets that contain 50, 100, 200, 500, and 1000 images (out of 4379 daylight hours) are selected using the K-means clustering algorithm. The validation set contains 10% randomly selected images from the training set and the test set contains 500 randomly selected images from the entire database. The results show that predicted image error, measured by $\mathcal{L}_{MSE}$ and $\mathcal{L}_{RER}$ with normalized luminance values, decreases with the number of training samples. However, the error curves decrease slowly after the number of images reaches 200. Therefore, 200 images are selected as the optimum sample size (5% of 4379 images) and the results are discussed in the following analysis.

### *Per-pixel Error*

Per-pixel errors are measured by $\mathcal{L}_{MSE}$ and $\mathcal{L}_{RER}$ used in the loss function. The errors are solid projected angle weighted and calculated with logarithmic ($\log_{10}$) luminance values to correlate with the human visual system (Fechner, 1966). The range of ground truth intensities is [0, 8.3]. The developed network succeeded in achieving a $\log_{10} \mathcal{L}_{MSE}$ of 9.54e-03 and a $\log_{10} \mathcal{L}_{RER}$ of 4.85e-02 on the test set.

Per-pixel errors do not necessarily prompt significant differences in lighting design evaluations. For example, if the sun is estimated to be located a pixel off its ground truth location, a significant $\mathcal{L}_{MSE}$ and $\mathcal{L}_{RER}$ error will incur but will hardly lead to a different design decision. Here, the per-pixel errors are provided as benchmarks for future studies. Metrics that are commonly used by design professionals, such as false-color images and daylight glare probability (DGP) (Wienold and Christoffersen, 2006) are utilized to evaluate the results. Spatial contrast metric RAMMG (Rizzi et al., 2004) is also used to compare the results.

### *False-color Image Comparison*

False-color images are used to study the luminance distributions and the resulting visual effect, comfort, and performance. Figure 9 illustrates the comparison results in false-color with a logarithmic scale, error maps denote the absolute difference between the ground truth and the predicted luminance maps. These are six representative samples under different sky conditions (clear, intermediate, and overcast) from the test set. The results show that: 1) The predicted panoramic luminance maps (generated from 5% sparse samples) are visually imperceptible from the ground truth results in false-color images in a 1-10,000 cd/m$^2$ range, without the utilization of error maps. This implies that the proposed method will lead to the same design decisions as conventional rendering methods, with orders of magnitude less calculation time. The error maps reveal that higher errors in each case occur in the window area. 2) Among all sky conditions, images of the clear sky with high direct irradiance values are most challenging to predict with the highest errors. This is mostly due to the higher luminance range in these images than those rendered under overcast sky conditions. Overall, the study obtains high-quality results illustrating the ability of the developed network to predict panoramic luminance maps under novel lighting conditions.

### *DGP*

DGP is a visual comfort indicator that describes the subjective magnitude of discomfort glare with the percentage of the population who would perceive the scene as glary (Wienold and Christoffersen, 2006). For DGP analysis, each panoramic luminance map is converted to 10 equidistant fisheye images over 360° of rotation in the y-axis using 36° increments. DGP is then applied to the predicted and the ground truth fisheye images. Figure 10 shows the per-view-angle DGP comparison of an exemplary test lighting condition (January 21$^{st}$ 11:30 am, direct and diffuse irradiances of 553 and 87 W/m$^2$), with error maps illustrating the absolute differences. The result illustrates that: 1) a subject's visual comfort level can change vastly across various view directions of the same scene depending on the shift of the sun location in and out of the field of view and the relative position of the sun in the field of view, 2) the neural network model accurately predicts the overall luminance environment, with the predicted DGP values from fisheye images closely matching those of ground truth ones in all directions, and 3) the more significant errors occur towards the window directions.

Figure 11 shows the comparison scatter plot of the DGP values for all of the test lighting conditions. It illustrates a relatively strong agreement (with an r$^2$ of 0.99 and MSE of 1.7e-4) with some noise at higher ranges. Further investigation shows the high errors occur around winter noon time, when sun direct appears in the field of view through the facing window (Figure 12). The model has difficulties in estimating the sun luminance (1.6 × 10$^{+9}$ cd/m$^2$) which is multiple orders of magnitude higher than the rest part in the scene. The failure cases refer to approximately 1% of the total tests cases, and the predictions still fall in the same zone (intolerable) as the ground truth DGPs. The situation is avoidable in the real world, when people have an ability to change their view directions and positions to avoid a direct view of the sun. For scenes with actual DGP in the imperceptible, perceptible and disturbing zone, DNN method produced very little variation.

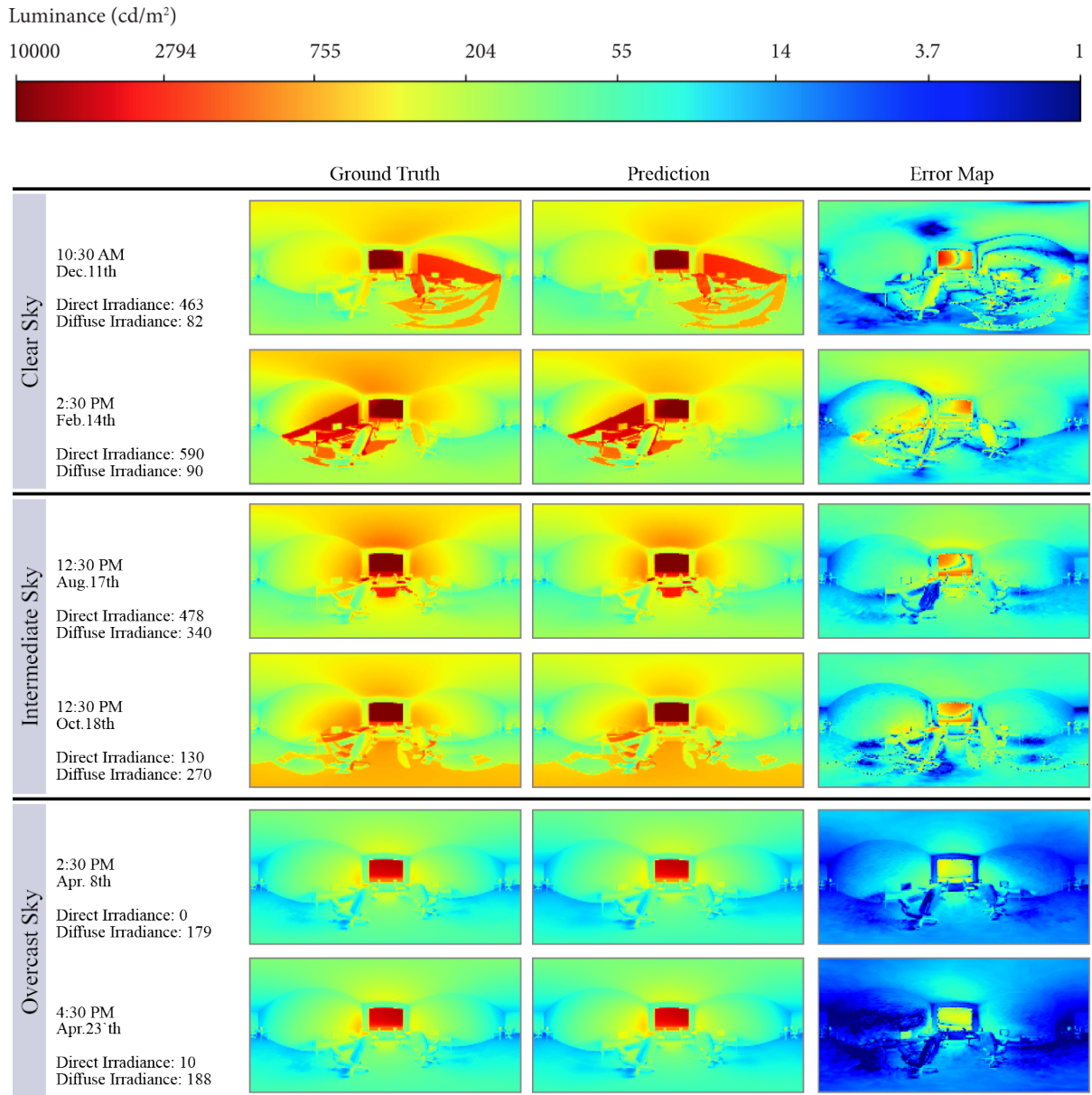

Figure 9. Six exemplary test cases with different sky types are displayed in false-color with a logarithmic scale.

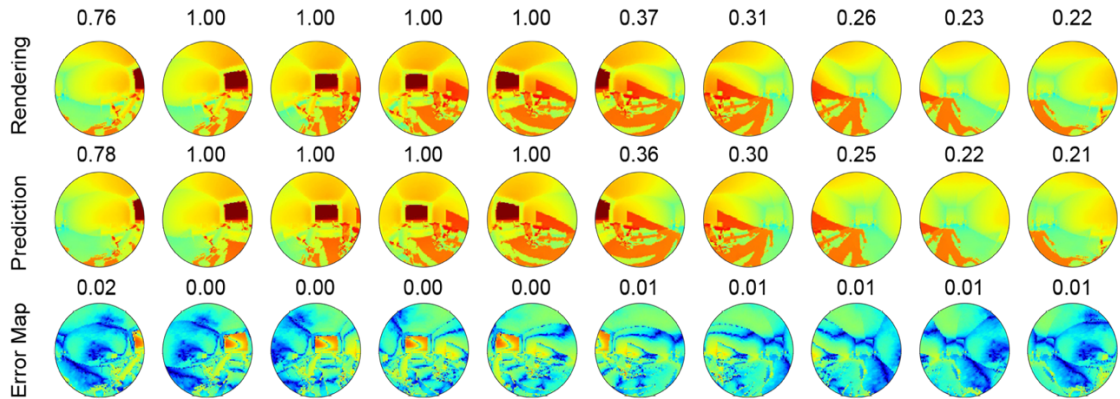

Figure 10. Comparison of Radiance rendered and DNN predicted fisheye images over a 360° view in 36° increments. The ground truth and predicted fisheye images are labeled with the DGP value. The error maps are labeled with absolute difference in DGP values. All images are shown in false-color with a logarithmic scale between 1-10,000 cd/m$^2$.

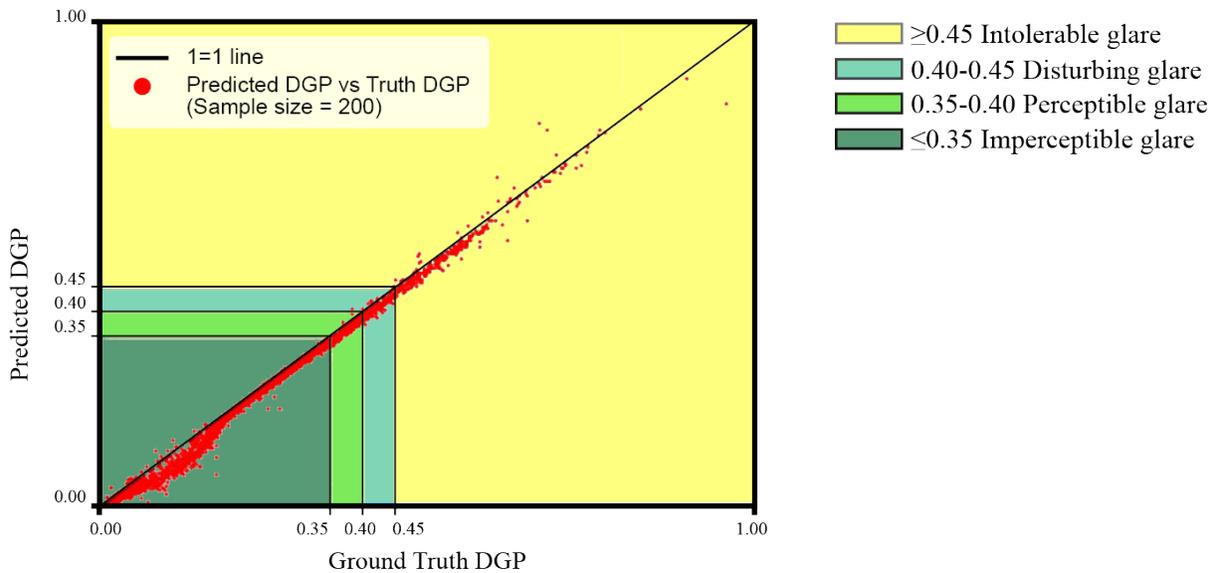

Figure 11. Scatterplot comparison of DGP values calculated using ground truth and predicted images

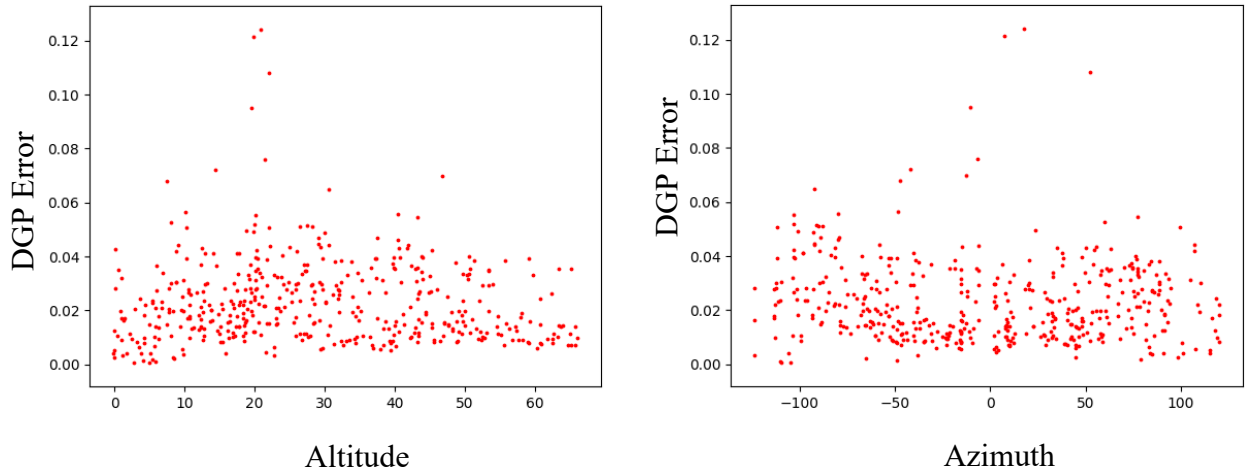

Figure 12. Scatter plots of DGP prediction errors (differences between predictions and ground truth) and the sun altitude (left) and azimuth (right). The max error occurs at sun altitude and azimuth of roughly 20° and 0°, respectively, when sun directly appears in the field of view through the south window in the Winter.

### *RAMMG*

RAMMG (Rizzi et al. 2004) is a spatial contrast measure of images and is utilized as a perceptual predictor of contrast-based visual characteristics of daylight (Rockcastle et al., 2017). RAMMG creates an image pyramid by sub-sampling an image to n-levels. At each level the resolution is reduced to half of the previous level. The RAMMG measure is the mean contrast calculated at each pyramid level. RAMMG is applied to both the predicted panoramas and Radiance rendered panoramas in the test set. Subsampling of five levels is used taking into consideration the image resolution (460 by 230) (Figure 13).  Figure 14 illustrates the correlation between the predicted and ground truth RAMMGs; it shows a strong agreement ($r^2$ of 0.98) with a few errors at higher ranges, when the sun directly appears through the window ($r^2$ of 0.99 if 4 outlier cases are excluded).

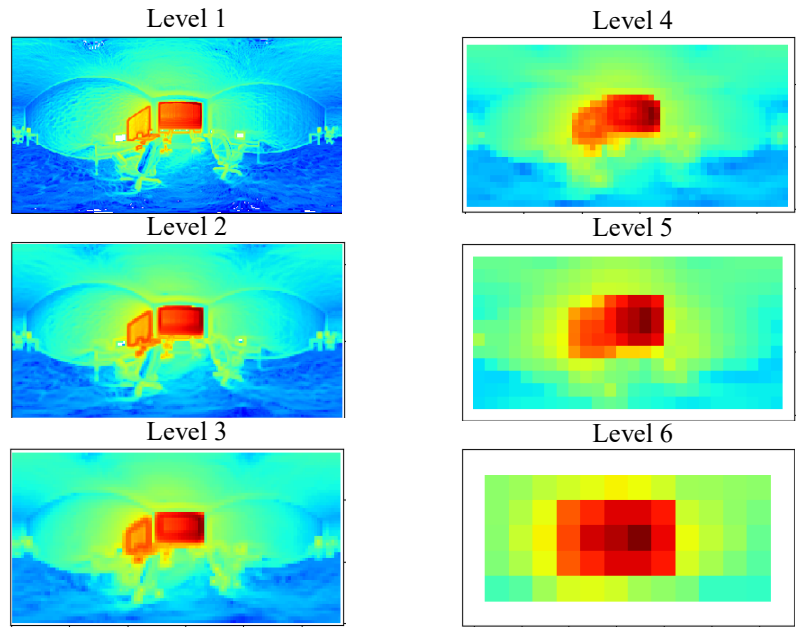

Figure 13. Each image represents a contrast map (each pixel represents the sum of neighborhood contrast of that pixel) of a sample panorama, calculated at every subsampling level starting from level 1. The image resolution at each level is reduced to half of the previous level. The contrast map starts providing less meaningful information from level 6.

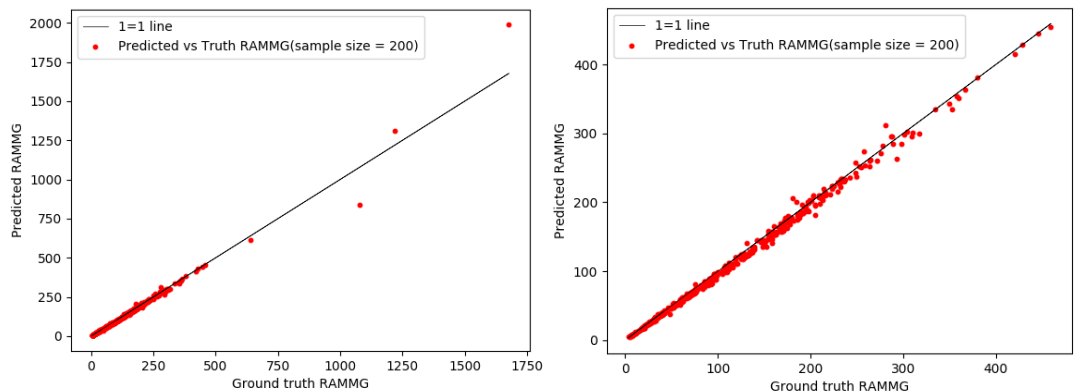

Figure 14. Comparison of RAMMG values calculated using ground truth and predicted images: (Left) original data, and (Right) zoomed in data.

## Optimizing the Data Collection Period

The DNN model can accurately predict annual panoramic luminance maps with only 5% of data selected using a general K-means clustering method (Training Set 1). However, the training set is evenly distributed over the entire year. Such a training set can be generated using simulation methods, but it is not feasible with in-situ data collection. Therefore, it is necessary to explore training with samples of a single continuous collection period. A sensitivity analysis is performed to find the minimum and optimum collection period, and the effect of different collection periods is tested on the prediction accuracy of the annual luminance maps. Longer collection periods, which cover more sun positions and different sky conditions are expected to produce more accurate results, but the added burden of elongated data collection periods may have diminishing returns. The objective is to find the shortest collection period that produces similar accuracy to full-year sampling. One-month data is selected, and the collection period is further reduced to two weeks and one week. The previous dataset (contains samples evenly distributed over the year) is used as the benchmark for comparisons (Training Set 1). The same machine learning workflow is performed with different training datasets to predict the annual luminance maps under various sky conditions.

### *Training Set 2: The training samples are collected from one-month continuous data*

Daylighting parameters for the one-month samples are illustrated in Figure 15 - Figure 17. Compared to the evenly distributed samples, one-month samples cover a small portion of the annual data. As a result, the model is trained with a limited amount of variability. Additionally, each month presents a different sky data, therefore presents different training opportunity. During equinoxes (March and September), the sun date lines spread to larger portions of the sky over the month (higher percentages of sky dome in Figure 15), which means the data from these months can better represent the variance of the sun position parameters. Conversely, during the winter and summer solstices (December and June), the date lines remain closer to each other (lower percentage of sky dome in Figure 15), which means the data from these months encapsulate the limited variance of the annual sun path diagram.

Figure 16 shows the distribution plots of direct and diffuse sky irradiances. The red, green, and blue dots represent samples from training, validation, and test sets, respectively. The red dots cover more sky conditions in summer compared to winter. There are not any red dots in the upper right zones in winter months, which indicates that the training/validation sets from these months do not contain samples with skies of high direct and diffuse irradiances. Figure 17 presents the joint probability distributions of training samples over the sky condition parameters. All diagrams have been processed as four by four blocks and normalized by the number of samples in the month, where the color of each block represents the percentage of samples in that range. The diagram shows that among all months, the block with the lowest diffuse and direct irradiances (bottom left corner) has the highest probability density (i.e., overcast skies are the dominant sky conditions in Seattle). Summer months (July, August, and September) have darker blue blocks with high direct and low diffuse irradiances. Sunny skies are more common in these months. The blocks cover a larger portion of the sky irradiances over summer months, which means the data from these months can better represent the variance of the sky condition parameters. This kind of analysis can be useful to select the periods of data generation and collection when working with other climatic locations.

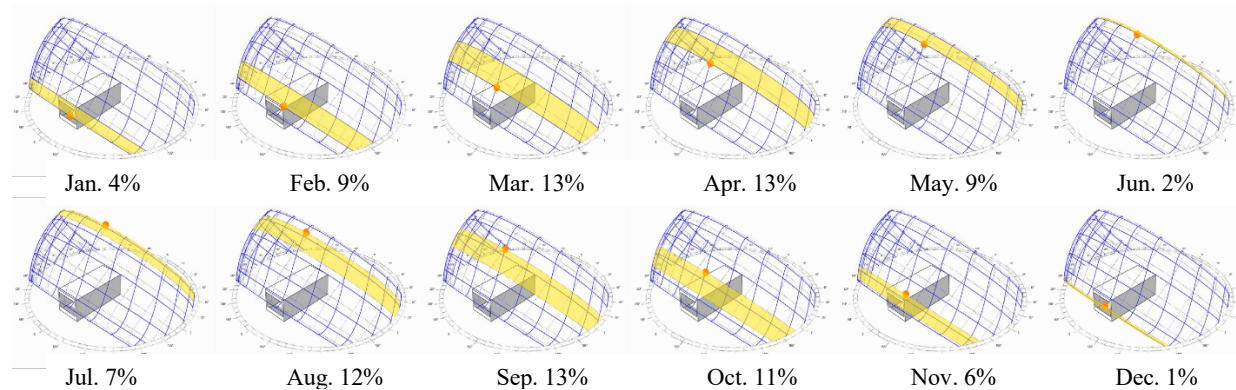

Figure 15. Sun path diagrams of Seattle with possible sun positions of each month shown in yellow, marked with the percentage of the sun area that each month covers (yellow area).

### *Comparisons of Predictions with Training Set 2*

The predictions are generated with one-month training samples and evaluated by per-pixel errors, visual discomfort metric DGP, and luminance contrast metric RAMMG. Figure 18 illustrates comparisons of the predicted luminance maps when trained with one of the twelve-month training sets and the training set 1. The differences are measured by all the above evaluation metrics. The results show: 1) Significant differences are found among predictions generated with different training sets; 2) There exists a pattern that lower prediction errors occur around the equinoxes, while highest errors occur around solstices, which implies that the most influential aspects on the prediction accuracies are the parameters of the sun positions; 3) Predictions generated with training samples from the summer months perform better than those from the winter months, indicating that the sky parameters play a secondary role.

To better illustrate the results, Figure 19 shows the DGP distribution comparisons among the predictions generated with one of the twelve-month training sets. The figures illustrate that: 1) similar to the previous analyses, the agreement between ground truth DGPs and the predicted DGPs generated with data selected around equinoxes is stronger than those generated with data selected around solstices; and 2) with training samples selected around equinoxes (e.g., March, April, September, and October), the predicted DGPs closely match to ground truth values.

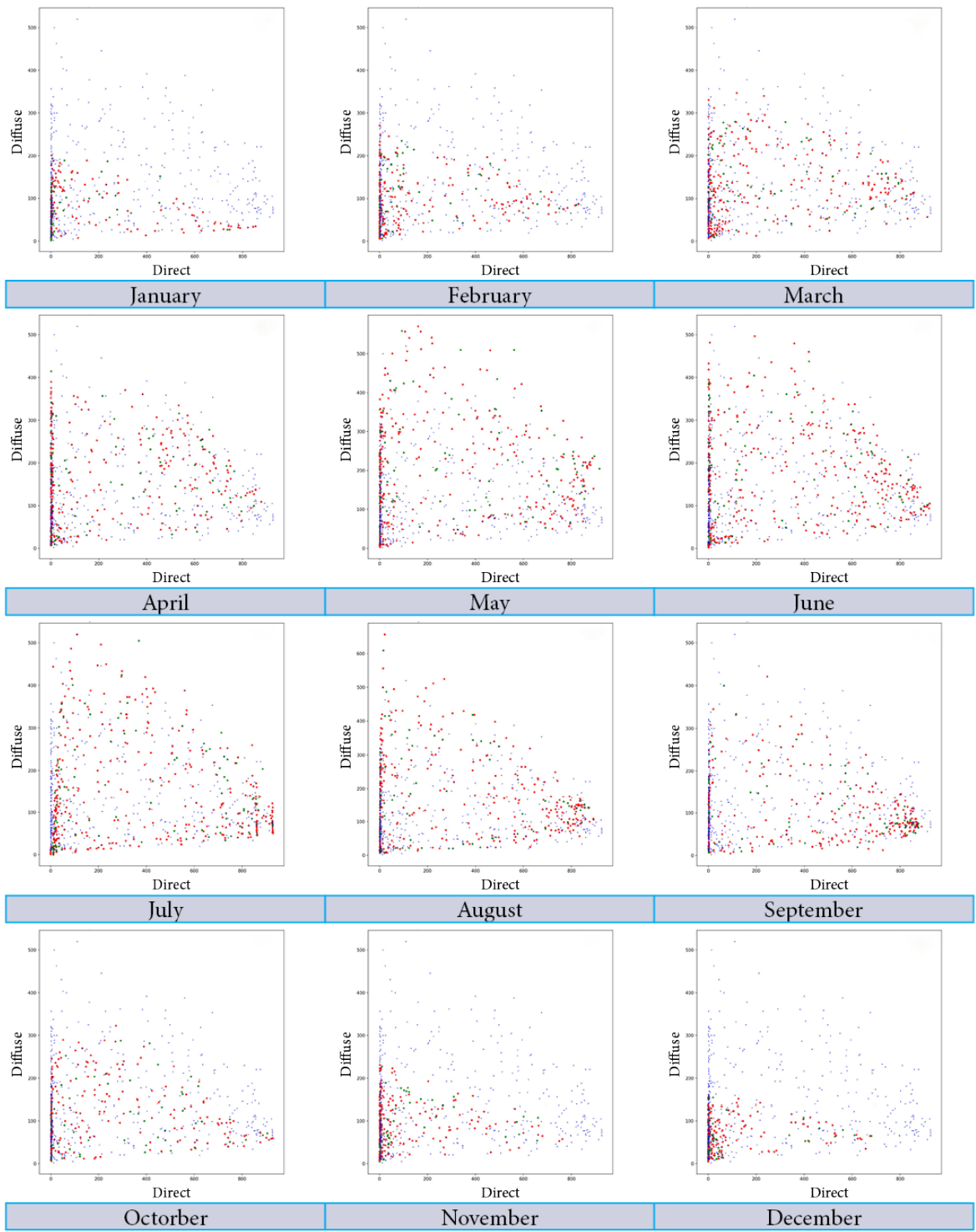

Figure 16. Distribution plots of selected samples over the sky condition parameters (direct and diffuse irradiances). The red, green, and blue dots represent samples from training, validation, and test sets. The red dots cover more sky conditions in summer compared to winter.

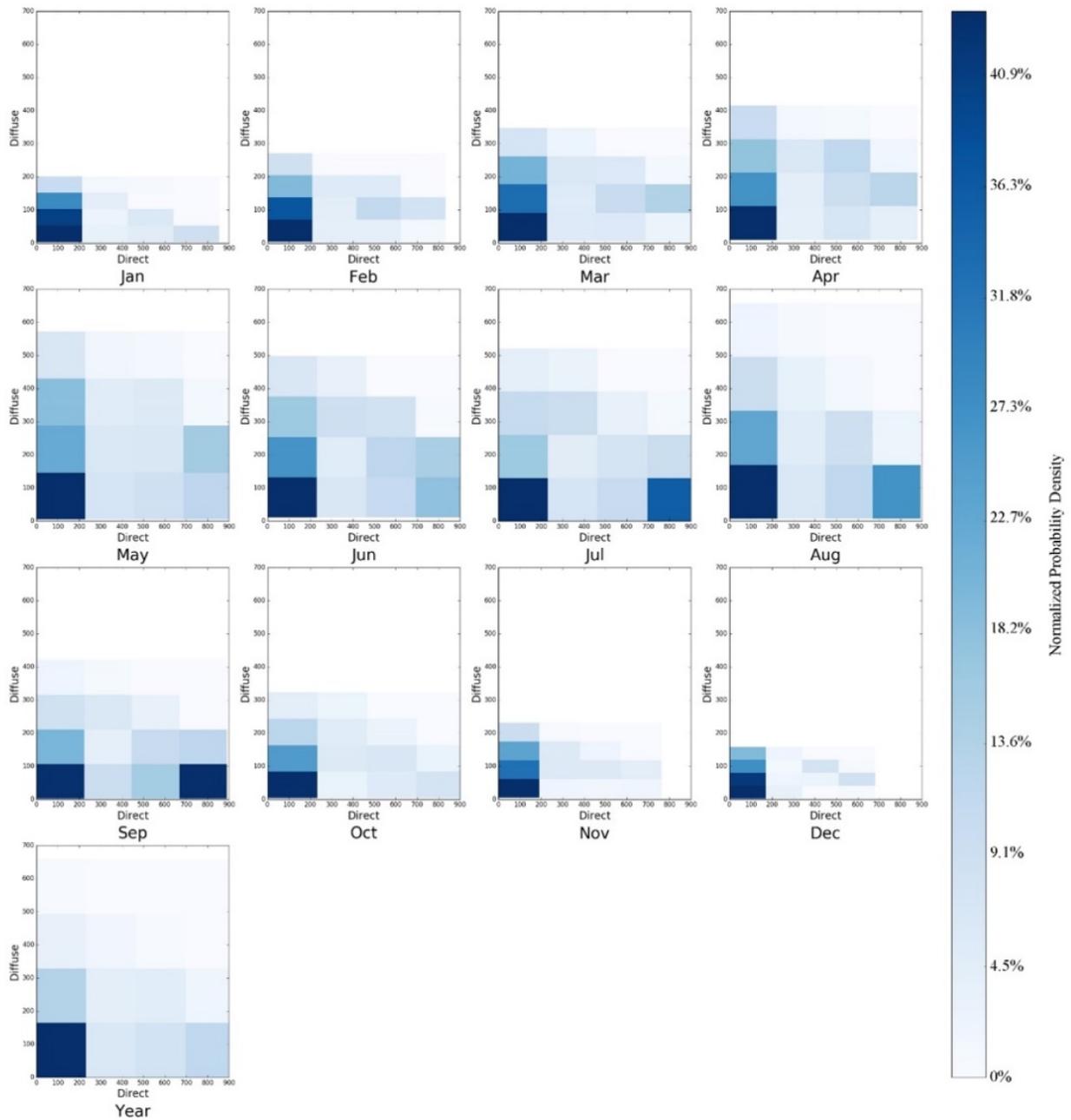

Figure 17. Joint probability distribution diagram of selected samples over the sky condition parameters (direct and diffuse irradiances).

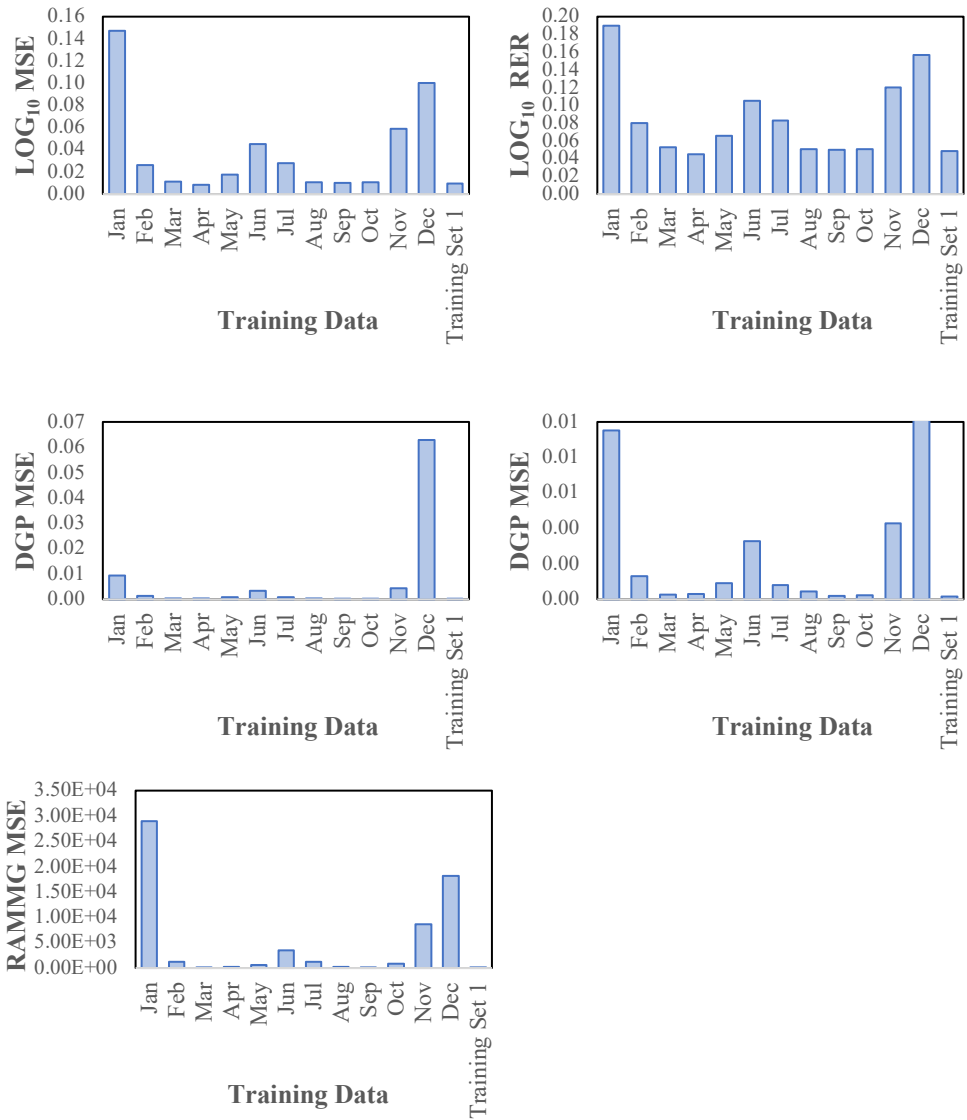

Figure 18. Comparisons of prediction results generated with monthly training samples and the training set 1, evaluated by: Log10 MSE errors (Top Left) and Log10 RER errors (Top Right); DGP MSEs - Original (Middle Left), -December column was truncated in order to make the whole distribution display better (Middle Right); RAMMG MSEs (Bottom).

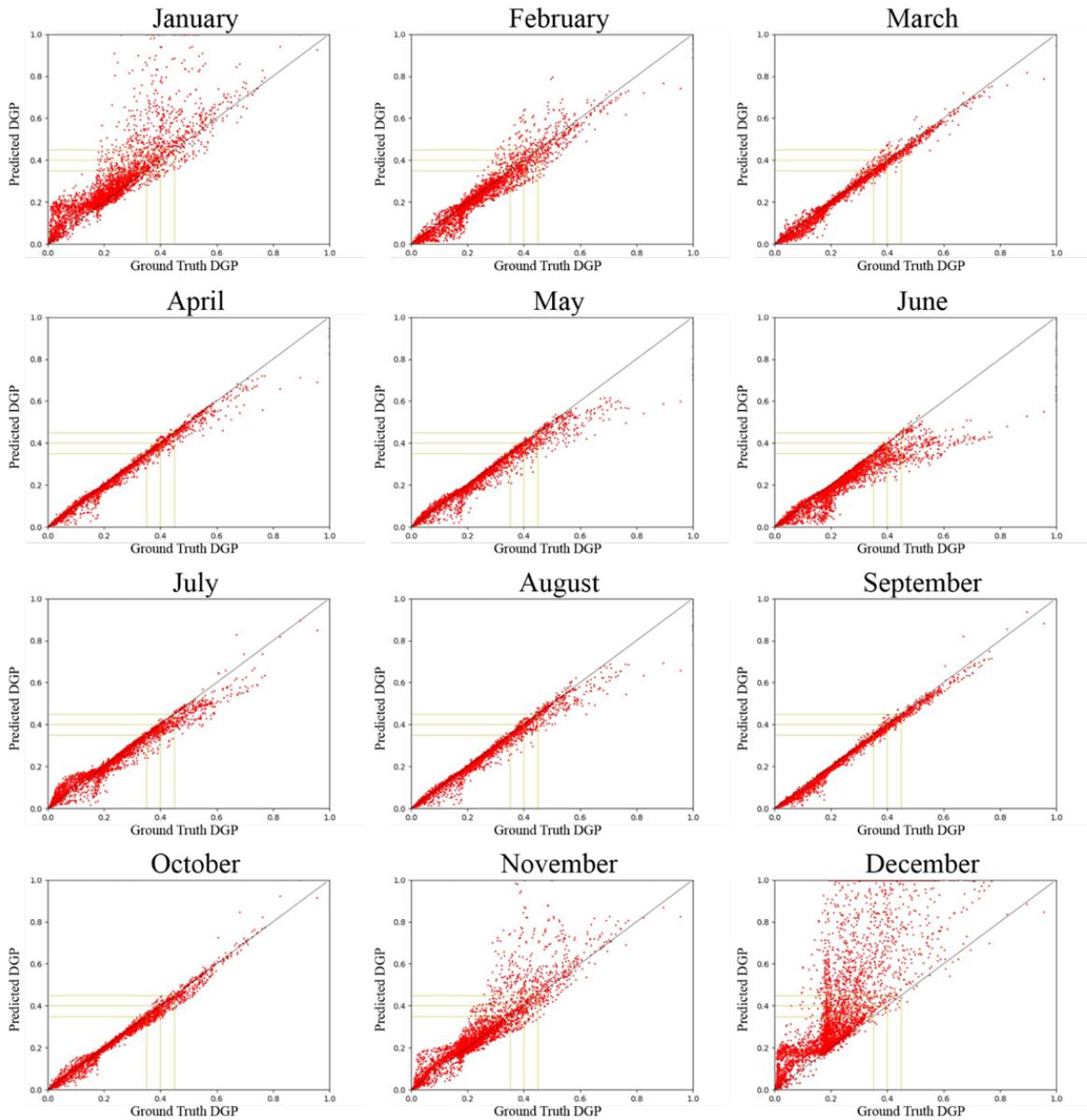

Figure 19. Scatter plots compare DGP values calculated using ground truth fisheye images and predicted fisheye images which are generated with monthly training samples.

The analyses show that it is possible to accurately predict the annual luminance maps, with one-month continuous data. The best months to capture or generate training data in Seattle are around equinoxes (March, April, September, or October, each month approximately corresponding to 8% of the entire year). The results also imply that a better strategy than evenly selecting samples over the light domain is adding more weights to the sun position parameters, as the sun positions have a more significant impact on the prediction accuracies than the sky conditions.

Datasets of shorter collection periods of continuous hourly 1- or 2-weeks are compared against the training sets of 1 and 2. The results show that using training sets of shorter collection periods (less than one month) will greatly reduce the prediction accuracies. Figure 20 shows the correlations between the predicted and ground truth DGPs for March. The results confirm that DGP predictions generated with the training sets of shorter collection periods have weaker agreements with the ground truth, with $r^2$ of 0.770, 0.945, and 0.990 for a training set of 1-week, 2-weeks, and 1-month data, respectively. DGP predictions generated with the training set of 1- or 2-weeks data have significant errors that occur at high ranges, indicating that the model has diminished capability in accurately predicting the high luminance or high contrast scenes.

Overall, the results show that the minimum continuous data collection period for accurate predictions is one month. Further reducing the data collection period to two weeks or a single week results in significant errors and may be misleading in a design process.

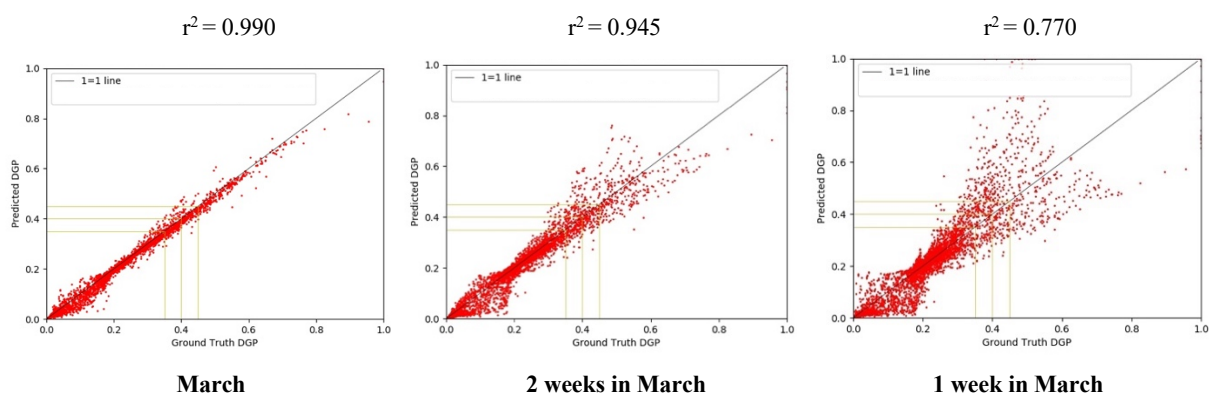

Figure 20. Scatter plots comparison between DGP values calculated using the predicted fisheye images generated with three training sets and the ground truth ones.

### Training Set 3: The training samples are collected from 9 days of hourly data collected around the spring equinox, summer and winter solstices

Training with data collected from multiple short periods is expected to have increased prediction accuracy compared to the training with data collected from a single continuous period, due to the expanded distribution of the sun locations and sky conditions. Although continuous data is more desirable in field studies, discrete data can easily be generated using the simulation tools. However, it is desirable to limit the number of simulations necessary to train the DNN. The training sets in this section contain samples collected from multiple short periods around the equinoxes and solstices. Equinoxes and solstices are selected for two reasons. Sun position variance is found to have the most significant impact on the prediction accuracies in previous analyses. Secondly, equinoxes and solstices are key time points, which architects and lighting professionals typically simulate. Therefore, using these samples to predict the year-round luminance maps will match the current practices of performing a limited number of point-in-time simulations. The training set 3a contains samples collected from 4 short periods, each period consists of 3 days around the spring and fall equinoxes, and summer and winter solstices. The training set 3b excludes the samples of the fall equinox but keeps those of the spring equinox. The duration of each period selected from equinoxes and solstices in the third training set 3c is reduced to one day. The total amount of samples in the training sets 3a, 3b, and 3c are 144, 108, and 48, respectively, all less than 200 samples in the benchmark training set (Training Set 1).

To illustrate the lighting characteristics of all the training sets, Figure 21 shows the distribution plots of selected samples over the four-dimensional lighting domain (sky direct and diffuse irradiances and sun azimuth and altitude angles).

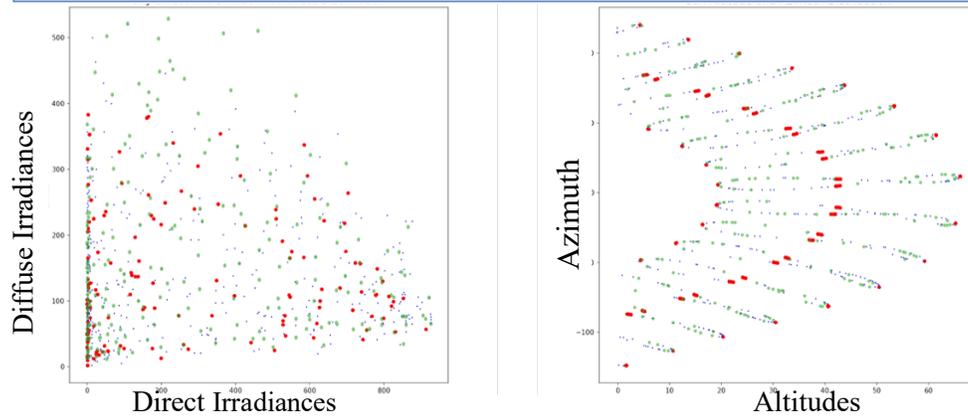
Training Set 3a: 12 Days

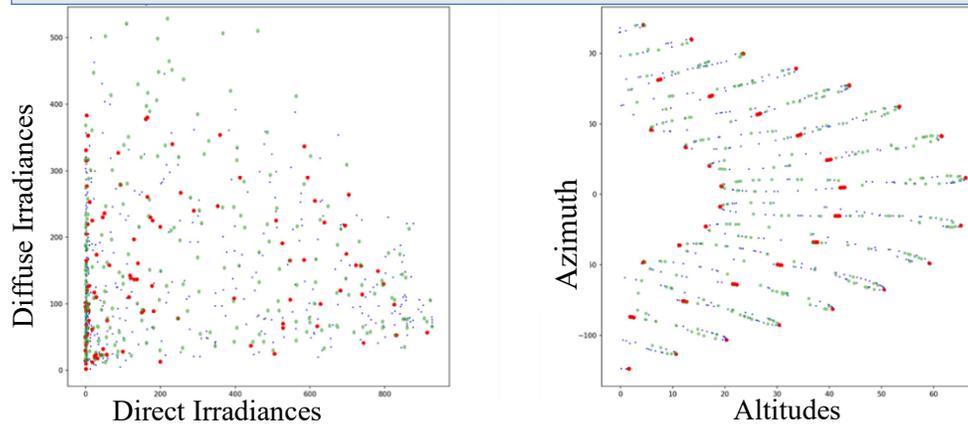
Training Set 3b: 9 Days

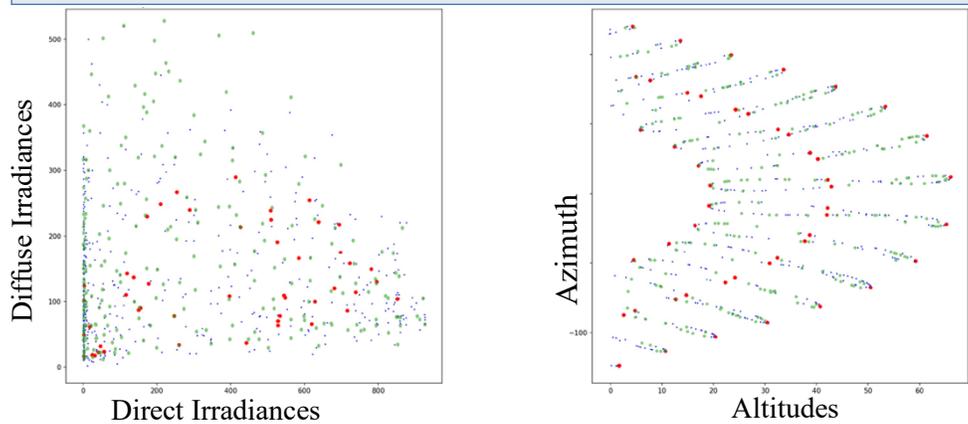
Training Set 3c: 4 Days

Figure 21. Distribution plots of selected samples over the four-dimensional lighting domain: (Left column) distribution plots over sky condition parameters (direct and diffuse irradiances), and (Right column) over sun position parameters (azimuth and altitude angles). Red dots represent samples in each training set while green dots represent samples of the benchmark dataset (Training Set 1).

### *Comparisons of Predictions with Training Sets 3a, 3b, and 3c*

In all four comparisons, predictions generated with the training set 3a containing 12-days of samples have the lowest errors, while those with the training set 3c containing 4-days of samples have the highest errors. As expected, the prediction errors increase with shorter data periods. Table 2 presents the comparisons between predictions generated with different training sets. The most important finding is that the training sets 3a and 3b have similar or even less per-pixel errors than the training set 1. This suggests that the method can have improved prediction accuracies with fewer training samples through better sample selection strategies.

Figure 22 illustrates the correlations between DGP values calculated using predicted and ground truth luminance maps. The results show strong ($r^2 = 0.99$ and $0.98$) agreements between predicted DGPs, generated with the training sets 3a and 3b, and the ground truth DGPs, with a few errors (1% of the total test cases) at higher ranges, when the sun direct appears in the field of view through the south-facing window. Overall, the training sets 3a with 12-days samples and 3b with 9-days samples succeed in generating luminance maps that are very close to ground truth in all view directions. Prediction results of the training set 3c with 4-days samples yield less satisfactory results. Given that the differences between training sets 3a and 3b are not significant, it is preferable to use the training set 3b that requires only 9 days of data (Figure 23) to predict the entire year.

Table 2. Comparison between predictions with different training sets

|  | 12 Days | 9 Days | 4 Days | Training Set 1 |
|---|---|---|---|---|
| **$LOG_{10}$ MSE** | 5.58E-03 | 7.99E-03 | 4.91E-02 | 9.54E-03 |
| **$LOG_{10}$ RER** | 3.71E-02 | 4.44E-02 | 1.10E-01 | 4.85E-02 |
| **DGP MSE** | 1.60E-04 | 3.12E-04 | 2.54E-03 | 1.74E-04 |
| **RAMMG MSE** | 6.60E+01 | 3.04E+02 | 6.40E+02 | 2.92E+01 |

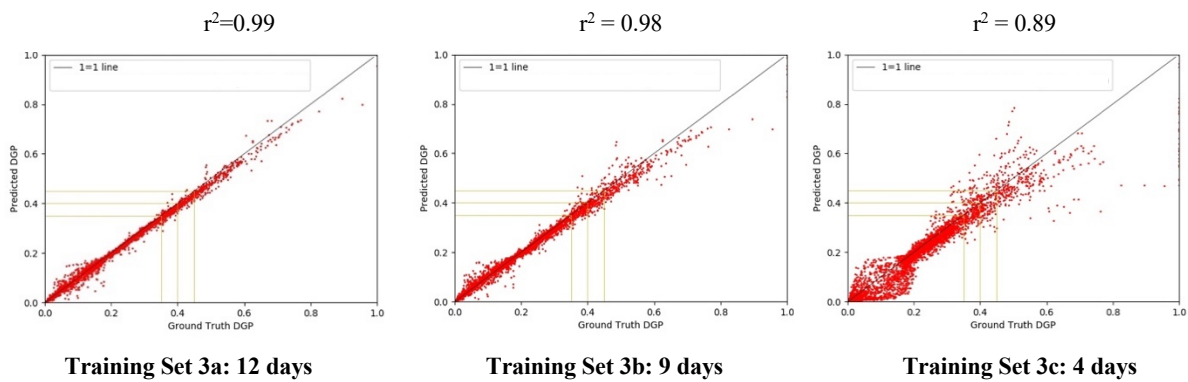

| Training Set 3a: 12 days | Training Set 3b: 9 days | Training Set 3c: 4 days |

Figure 22. Scatterplots show the correlation of DGP values calculated using ground truth and predicted fisheye images. Red dots represent DGP pairs.

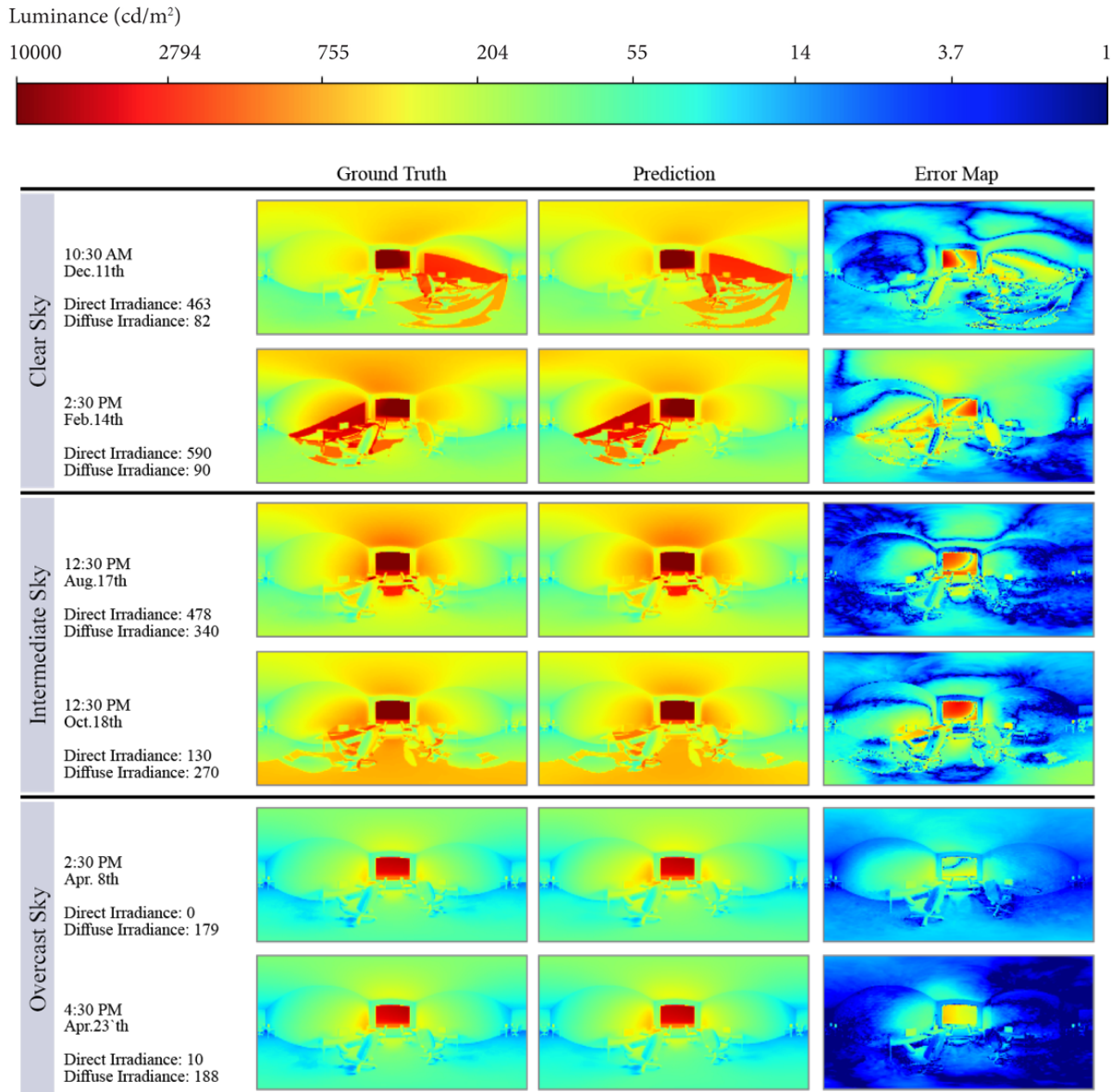

Figure 23. Six representative cases under different sky conditions, generated with the training set 3b containing 9-days samples, are displayed in false-color with a logarithmic scale.

## Conclusion

This paper demonstrates the utilization of a DNN approach for predicting annual panoramic luminance maps of indoor spaces under varying sun and sky conditions from a limited number of HDR images. Annual panoramic luminance maps can be faithfully predicted with 2.5% of annual renderings from three short periods (9 days in total), and only requires 30 minutes of computing time during training, and a few seconds during evaluation. Annual luminance-based simulation using this method is magnitudes faster (~40 times speedup) than conventional simulation methods with comparable accuracy examined by per-pixel errors, false-color image techniques, visual comfort, and contrast metrics.

In general, practitioners are less motivated to integrate annual luminance-based metrics into the design evaluation process due to the time-consuming rendering pro[1]cesses. Making annual luminance-based simulations more accessible will enable further utilization of luminance-based metrics in practice. The availability of the data among researchers will promote the development of new luminance-based metrics. Panoramic luminance maps (with 360° horizontal and 180° vertical field of view) are recommended, as they provide the ability to evaluate a subject's visual experience over multiple viewing directions.

The DNN model separates the direct and indirect contributions from the sun to reconstruct sharp shadows and accurate sun penetrations. The findings reveal that compared to sky conditions, sun position parameters play a more important role in generating accurate predictions. This is useful information in refining the sampling strategies for future explorations.

A public dataset of the high-quality annual HDR panoramic luminance maps used in this study and the code is available (https://github.com/yueAUW/neural-daylighting).